\documentclass[letterpaper]{article} 
\usepackage{aaai2026}  
\usepackage{times}  
\usepackage{helvet}  
\usepackage{courier}  
\usepackage[hyphens]{url}  
\usepackage{graphicx} 
\usepackage{natbib}  
\usepackage{caption} 
\usepackage{caption}

\usepackage{booktabs}
\usepackage{multirow}
\usepackage{amsmath}
\DeclareMathOperator*{\argmax}{arg\,max}
\usepackage{graphicx} 
\usepackage{makecell}
\usepackage{enumitem} 
\usepackage{amssymb}
\usepackage{mathtools}
\usepackage{float} 
\usepackage[ruled,vlined]{algorithm2e}
\usepackage{tcolorbox}
\usepackage[x11names]{xcolor}
\usepackage{listings}

\usepackage{}
\tcbuselibrary{raster}
\usepackage{algorithmic}

\usepackage{newfloat}
\usepackage{listings}
\DeclareCaptionStyle{ruled}{labelfont=normalfont,labelsep=colon,strut=off} 
\floatstyle{ruled}
\newfloat{listing}{tb}{lst}{}
\floatname{listing}{Listing}
\title{AgentSwift: Efficient LLM Agent Design via Value-guided Hierarchical Search}
\author{
    Yu Li\textsuperscript{\rm 1}, Lehui Li\textsuperscript{\rm 3}, Zhihao Wu\textsuperscript{\rm 2}, Qingmin Liao\textsuperscript{\rm 1}, Jianye Hao\textsuperscript{\rm 2}, Kun Shao\textsuperscript{\rm 2}, Fengli Xu\textsuperscript{\rm 1}\thanks{Corresponding author.}
}
\affiliations{
    \textsuperscript{\rm 1}Tsinghua University\\
    \textsuperscript{\rm 2}Huawei Noah’s Ark Lab\\
    \textsuperscript{\rm 3}Shandong University\\

    liyu24@mails.tsinghua.edu.cn, \{fenglixu, liaoqm\}@tsinghua.edu.cn, \{shaokun2, haojianye\}@huawei.com, 202200300096@mail.sdu.edu.cn
}

\usepackage{bibentry}

\begin{document}

\maketitle

\begin{abstract}

Large language model (LLM) agents have demonstrated strong capabilities across diverse domains, yet automated agent design remains a significant challenge. Current automated agent design approaches are often constrained by limited search spaces that primarily optimize workflows but fail to integrate crucial human-designed components like memory, planning, and tool use. Furthermore, these methods are hampered by high evaluation costs, as evaluating even a single new agent on a benchmark can require tens of dollars. The difficulty of this exploration is further exacerbated by inefficient search strategies that struggle to navigate the large design space effectively, making the discovery of novel agents a slow and resource-intensive process. To address these challenges, we propose AgentSwift, a novel framework for automated agent design. We formalize a hierarchical search space that jointly models agentic workflow and composable functional components. This structure moves beyond optimizing workflows alone by co-optimizing functional components, which enables the discovery of more complex and effective agent architectures. To make exploration within this expansive space feasible, we mitigate high evaluation costs by training a value model on a high-quality dataset, generated via a novel strategy combining combinatorial coverage and balanced Bayesian sampling for low-cost evaluation. Guiding the entire process is a hierarchical Monte Carlo Tree Search (MCTS) strategy, which is informed by uncertainty to efficiently navigate the search space. Evaluated across a comprehensive set of seven benchmarks spanning embodied, math, web, tool, and game domains, AgentSwift discovers agents that achieve an average performance gain of 8.34\% over both existing automated agent search methods and manually designed agents. Moreover, our framework exhibits steeper and more stable search trajectories. By enabling the efficient, automated composition of workflow with functional components, AgentSwift provides a scalable methodology to explore complex agent designs. Our framework serves as a launchpad for researchers to rapidly prototype and discover powerful agent architectures without the impediment of prohibitive evaluation costs.


\end{abstract}

\begin{links}
    \link{Code}{https://github.com/Ericccc02/AgentSwift}
\end{links}

\section{Introduction}
The recent rise of large language models (LLMs)~\cite{brown2020language,radford2018improving,radford2019language} has sparked an explosion of interest in agentic systems. Early forms of such systems, like Chain-of-Thought~\cite{wei2022chain}, Tree-of-Thought~\cite{yao2023tree}, Debate~\cite{du2023improving} and Self-Refine~\cite{madaan2023self}, exemplify the agentic workflow paradigm. These agentic workflows significantly boosted performance on reasoning-intensive tasks, such as mathematical problem~\cite{romera2024mathematical} and logical deduction~\cite{shang2024defint}. Subsequently, more advanced systems like Voyager~\cite{wang2024voyager} and AutoAct~\cite{qiao2024autoact} incorporated structured components such as planning, tool use, and memory. These functional enhancements allowed agents to handle a broader range of tasks—such as web interaction~\cite{nakano2021webgpt}, open-ended exploration~\cite{wang2024voyager}, and planning~\cite{xie2024travelplanner}—further extending their capability. These developments highlight the importance of agent design, yet building high-performing agents remains manual and labor-intensive, motivating the need for automated agent search.


Despite recent progress, the design of agentic systems remains largely manual and heuristic. Early efforts focused on prompt optimization~\cite{yang2024large,khattab2023dspy} or agent profiling~\cite{yuan2024evoagent}, while graph-based approaches~\cite{zhugegptswarm,zhang2024g} explored communication topology. These methods typically target isolated subsystems such as prompts, roles, or message flow. More recent works like AFlow~\cite{zhang2024aflow}, ADAS~\cite{hu2024automated}, and AgentSquare~\cite{shang2024agentsquare} formulate agent design as a search problem over agentic workflows, aiming to discover effective end-to-end configurations. While these advances mark a shift toward agent search, the search of agent remains inefficient.



This inefficiency stems from three major challenges. First, there is an under-exploitation of proven human designs: most existing methods restrict search to specific parts of the agent, such as prompts, profiles, or workflows. As a result, they fail to incorporate or discover critical functional components like planning, tool use, and memory—elements essential for building agents capable of tackling complex, multi-stage tasks. Second, the evaluation cost of agent search remains prohibitively high. According to AgentSquare~\cite{shang2024agentsquare}, evaluating a simple CoT agent based on GPT-4o in ALFWorld~\cite{shridhar2021alfworld} requires around \$60. In most existing methods, each newly generated agent must be fully evaluated on benchmark tasks to obtain feedback. This results in a large number of unnecessary evaluations for poorly performing agents, leading to wasted computation and prolonged search cycles. Third, in large design spaces, search efficiency suffers. While methods like AFlow and ADAS aim to optimize entire workflows based on performance histories, they often employ search strategies that explore the vast design space inefficiently. Addressing these limitations is crucial to unlocking the full potential of agentic system search.

\begin{figure*}[!t]
\begin{center}
\captionsetup{belowskip=1pt}
\includegraphics[width=0.7\textwidth]{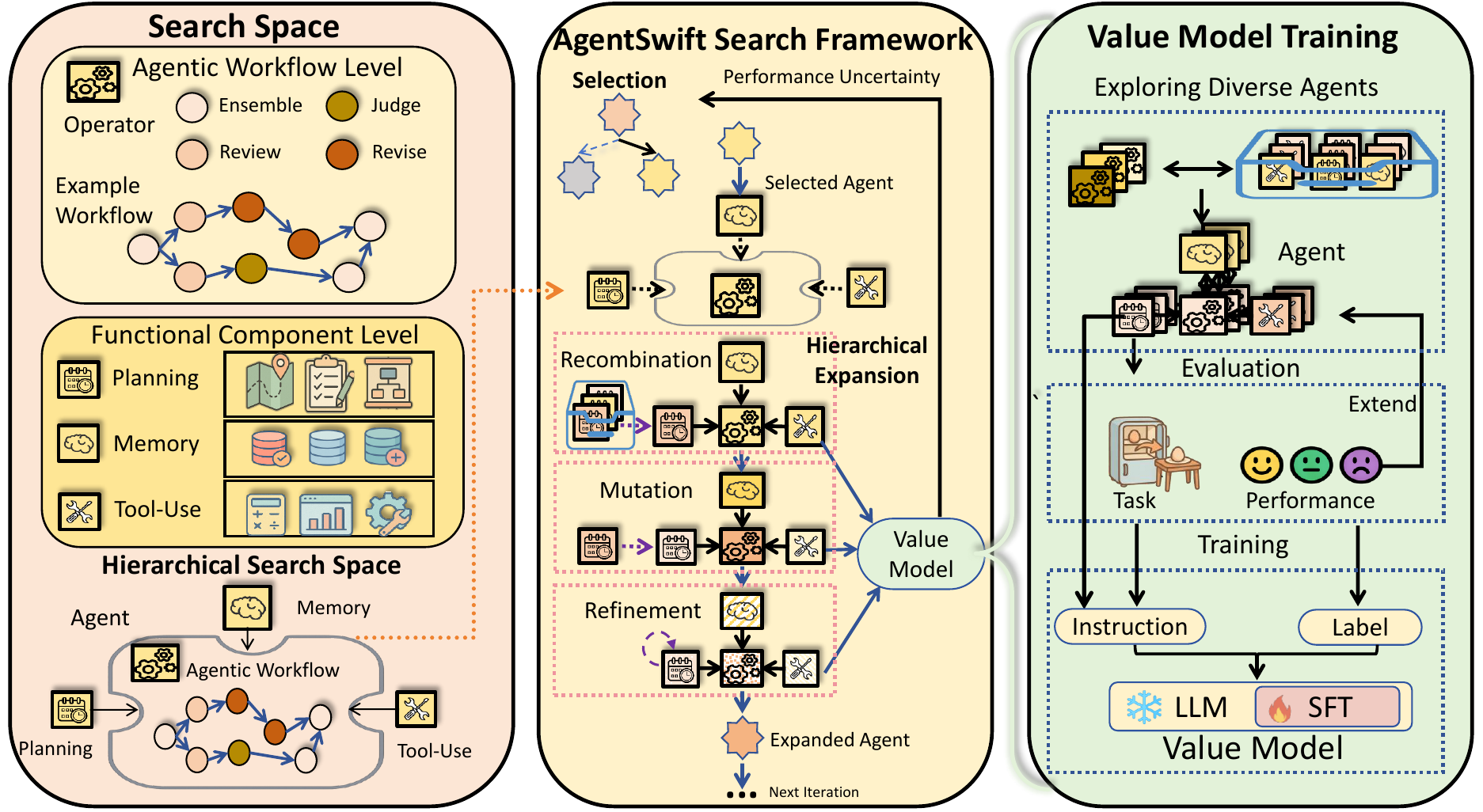}
\caption{Overview of our framework. The framework integrates (a) hierarchical search space (b) uncertainty-guided MCTS with hierarchical expansion (c) value model training}
\label{figure:main}
\end{center}
\end{figure*}

In this work, we propose a comprehensive framework that addresses these inefficiencies through three key innovations. \textbf{First}, we construct a hierarchical search space that includes both the agentic workflow and three functional components—\emph{memory}, \emph{tool use}, and \emph{planning}—that can be modularly attached to the agentic workflow. This search space extends the formulation of AFlow, enabling richer design possibilities beyond fixed workflow structures. This structured design space not only broadens the range of agent designs but also facilitates more meaningful performance modeling, making it well-suited for learning a predictive model. \textbf{Second}, we develop a value model that predicts the performance of a candidate agent given its design and a task description. To support effective learning, we construct a high-quality training dataset by combining pairwise covering arrays, which ensure comprehensive coverage of interactions between workflows and components, with balanced Bayesian sampling, which selects agent candidates from both high- and low-performing regions of the search space. This enables the model to generalize across a broad design space and provide accurate, low-cost predictions, effectively guiding the search process while avoiding unnecessary real-world evaluations. \textbf{Third}, we design an uncertainty-guided hierarchical expansion strategy based on Monte Carlo Tree Search (MCTS). During the MCTS expansion phase, the agent is iteratively improved through three operations—\emph{recombination}, \emph{mutation}, and \emph{refinement}—applied hierarchically to both the agentic workflow and functional components. In the recombination step, new candidates are sampled from a library of possible workflow structures or component implementations to replace existing ones. Mutation explores new candidates based on existing components and workflows, guided by the performance of previously evaluated agents. Refinement adjusts the agentic workflow and components based on feedback from failure cases. These modifications are guided by the value model’s predicted performance, ensuring the search explores promising directions efficiently. By comparing predicted and actual performance, we obtain a natural measure of uncertainty, which is integrated into the MCTS selection strategy to guide which to expand during the search. This integration of predictive modeling and uncertainty allows us to prioritize promising agent candidates, avoid unproductive regions, and conduct more targeted, efficient exploration of the design space. The overview of this work is illustrated in Figure ~\ref{figure:main}.

We validate our framework across seven widely-used benchmark datasets spanning domains such as math, web, tool, and game. Experimental results show that our method achieves an average performance improvement of 8.34\% over state-of-the-art baselines. The discovered agents generalize well across LLM backbones, demonstrating strong model-agnosticity. Additionally, our approach exhibits a steeper search trajectory, discovering high-performing agent designs with significantly fewer agent evaluations. Beyond final performance, our value model demonstrates high predictive accuracy and strong transferability to unseen tasks with minimal fine-tuning.

The key contributions of this work are as follows:
\begin{itemize}[leftmargin=*]
    \item We formalize the agentic system optimization as a hierarchical search over agentic workflow and functional components, establishing a general framework that extends prior approaches.
    \item We train a value model that predicts agent performance from agentic system and task description, enabling low-cost, model-driven evaluation during the search process.
    \item We propose an uncertainty-guided hierarchical expansion strategy based on MCTS, incorporating recombination, mutation, and refinement steps over both workflow and components.
    \item We empirically demonstrate the effectiveness of our method on seven diverse benchmarks, showing consistent improvements over state-of-the-art baselines.
\end{itemize}


\section{Related work}
\label{gen_inst}

\subsection{LLM agent}

Recent advances in LLM agents have introduced diverse agentic workflows that support multi-step reasoning via reflection and debate~\cite{wei2022chain,madaan2023self,du2023improving}. These workflows are often complemented by functional components that extend agent capability: memory supports long-term coherence and retrieval~\cite{wang2024voyager,wen2024dilu,park2023generative}, tool use enable interaction with external APIs~\cite{schick2023toolformer,qin2023toolllm,du2024anytool}, and planning facilitates subgoal decomposition and control~\cite{ge2024openagi,wang2024voyager,shen2023hugginggpt}. However, most agents are still manually designed for specific tasks, lacking a
unified framework that can systematically search and optimize across workflow and component
design choices.


\subsection{Automated agentic workflow design}

Early work on automating agentic workflows has largely focused on optimizing specific subsystems such as prompts~\cite{yang2024large,fernando2023promptbreeder}, agent profiles~\cite{chen2023autoagents,chen2023agentverse}, and communication topologies~\cite{zhugegptswarm,qian2024scaling,niu2025flow}. While these methods improve local components, they do not consider the agentic workflow as a whole. More recent approaches have attempted end-to-end agentic workflow search~\cite{zhang2024aflow,hu2024automated,shang2024agentsquare,zhang2025evoflow}. Extending this direction, MaAS~\cite{zhang2025multi} shifts from searching for a single optimal workflow to learning a query-conditioned distribution over agentic architectures, enabling adaptive deployment. However, these approaches still operate within predefined workflow primitives and often overlook the broader agent design space that includes functional components, limiting adaptability and extensibility.

\subsection{Performance predictor in AutoML}

The development of performance predictors in Neural Architecture Search (NAS) provides a valuable blueprint for progress in agentic system search. Early NAS efforts primarily focused on optimization strategies~\citep{zoph2016neural, real2019regularized, maziarz2018evolutionary}. While effective, these methods required costly evaluations of many candidate architectures. To overcome this limitation, the NAS community gradually introduced performance predictors~\citep{kandasamy2018neural, white2021bananas, qin2025transferrable}. This shift in NAS, from pure search to search guided by learned predictors, has led to significant improvements in efficiency. Notably, the research paradigm in NAS is closely aligned with the goals of agentic system design, as both involve navigating large design spaces under expensive evaluation constraints. Motivated by this connection, we incorporate a value model into agentic system search, enabling performance prediction for candidate agents and guiding the search process more efficiently.



\begin{figure*}[!t]
\begin{center}
\captionsetup{belowskip=-2pt}
\includegraphics[width=0.8\textwidth]{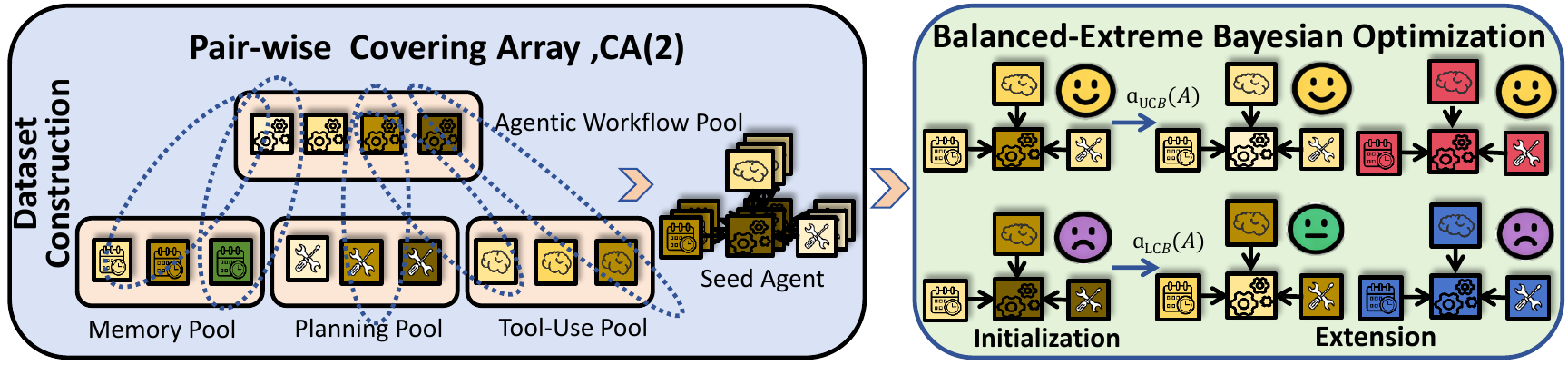}
\caption{Overview of dataset construction}
\label{figure:data}
\end{center}
\end{figure*}

\section{Search space}

More recent efforts like AFlow~\cite{zhang2024aflow} and ADAS~\cite{hu2024automated} treat the agentic workflow as a whole and perform end-to-end search over its structure. Despite their broader scope, these methods do not support flexible integration of functional components such as memory, planning, or tool use. Although AgentSquare~\cite{shang2024agentsquare} introduces these components into its design space, they are combined under a fixed agentic workflow template with rigid interfaces, and its search process remains prompt-centric. In contrast, we propose a hierarchical search space that jointly explores both the agentic workflow and composable functional components. 

\subsection{Agentic workflow}
Following AFlow~\cite{zhang2024aflow}, we define an agentic workflow $\mathbf{W}$ as a series of LLM-invoking nodes connected by edges to specify execution order. Formally, an agentic workflow $\mathbf{W}$ consists of a set of nodes $N$ and a set of edges $E$, written as $\mathbf{W} = (N, E)$. Each node $N_i \in N$ represents a single execution step and is characterized by the following parameters:
\begin{equation}
    N_i = (M_i,\;P_i,\;\tau_i,\;F_i),
\end{equation}
where $M_i \in \mathcal{M}$ is the language model used at this node, $P_i \in \mathcal{P}$ is the prompt provided to the model, $\tau_i \in \mathcal{T}$ is the decoding temperature, and $F_i \in \mathcal{F}$ specifies the output format. The edges $E \subseteq N \times N$ define the control and data flow between nodes, specifying the execution order. 


The agentic workflow search space is defined as:
\begin{equation}
\begin{split}
    \mathcal{S}_{\text{workflow}} = \big\{ (N, E) \;\big|\; & N_i = (M_i, P_i, \tau_i, F_i), \\
                                          & M_i \in \mathcal{M},\; P_i \in \mathcal{P}, \; \tau_i \in \mathcal{T}, \\
                                          & F_i \in \mathcal{F},\; E \subseteq N \times N \big\}.
\end{split}
\end{equation}

\subsection{Functional components}

In addition to the agentic workflow, we extend the search space to include composable functional components that provide essential agentic capabilities. Specifically, we consider three component types: \emph{memory}, \emph{tool use}, and \emph{planning}. These components are designed to be plug-and-play and can be integrated at specific points within the agentic workflow—for instance, a memory component may interact with a node to retrieve or store context, tool use can augment a node with external API calls, and planning can precede downstream execution steps.  


 \textbf{Memory.} The memory component allows agents to retrieve and incorporate information. It is defined as \( \mathbf{M} = (m,\;\tau,\;d) \), where \( m \) is the prompt used to query or update memory, \( \tau \) is the decoding temperature for memory-related LLM calls, and \( d \) denotes the external memory backend, such as a vector database.

\textbf{Tool Use.} The tool use component enables the agent to interact with external APIs or environments. It is defined as \( \mathbf{T} = (t,\;\tau,\;u) \), where \( t \) is the tool invocation prompt, \( \tau \) is the decoding temperature, and \( u \) represents the accessible toolset.

\textbf{Planning.} The planning component supports task decomposition and hierarchical control. It is defined as \( \mathbf{P} = (p,\;\tau) \), where \( p \) is the prompt for generating subgoals or plans, and \( \tau \) is the temperature used during plan generation.





The component search spaces are defined as:
\begin{equation}
\begin{aligned}
    \mathcal{S}_{\text{memory}} &= \left\{ (m,\; \tau,\; d) \;\middle|\; m \in \mathcal{P},\; \tau \in \mathcal{T},\; d \in \mathcal{D} \right\}, \\
    \mathcal{S}_{\text{tool}} &= \left\{ (t,\; \tau,\; u) \;\middle|\; t \in \mathcal{P},\; \tau \in \mathcal{T},\; u \in \mathcal{U} \right\}, \\
    \mathcal{S}_{\text{planning}} &= \left\{ (p,\; \tau) \;\middle|\; p \in \mathcal{P},\; \tau \in \mathcal{T} \right\}.
\end{aligned}
\end{equation}

Here, $\mathcal{P}$ is the prompt space, $\mathcal{T}$ is the temperature space, $\mathcal{D}$ is the space of memory backends, and $\mathcal{U}$ is the space of available tools.
\subsection{Hierarchical search space}

We define an agent $\mathbf{A}$ as a combination of an agentic workflow and a set of functional components. Formally, the agent is represented as:
\begin{equation}
    \mathbf{A} = (\mathbf{W},\; \mathbf{M},\; \mathbf{T},\; \mathbf{P}).
\end{equation}

The full agent search space is given by:
\begin{equation}
\begin{split}
    \mathcal{S}_{\text{agent}} = \big\{\mathbf{W},\;\mathbf{M},\;\mathbf{T},\;\mathbf{P} \;\big|\;
                                        & \mathbf{W} \in \mathcal{S}_{\text{workflow}},\mathbf{M} \in \mathcal{S}_{\text{memory}}, \\
                                        & \mathbf{T} \in \mathcal{S}_{\text{tool}},\mathbf{P} \in \mathcal{S}_{\text{planning}} \big\}.
\end{split}
\end{equation}
This formulation defines a hierarchical search space where both the structure of the agentic workflow and the configurations of its functional components are jointly optimized, enabling flexible composition, deeper architectural variations, and the reuse of classical human-designed modules. It subsumes existing methods such as AFlow~\cite{zhang2024aflow} and AgentSquare~\cite{shang2024agentsquare} as special cases within a more expressive and extensible design space.





\section{AgentSwift framework}
\label{Our search framework}
\subsection{Overview}

Given a task description $d$ and a performance evaluation function $\text{Eval}_d(\cdot)$, our objective is to find the agent design $\mathbf{A}^*$ from the joint search space $\mathcal{S}_{\text{agent}}$ that maximizes expected task performance. The optimization problem is defined as:
\begin{equation}
\mathbf{A}^* = \argmax_{\mathbf{A} \in \mathcal{S}_{\text{agent}}} \text{Eval}_d(\mathbf{A}) = \argmax_{(\mathbf{W}, \mathbf{P}, \mathbf{T}, \mathbf{M})} \text{Eval}_d(\mathbf{W}, \mathbf{P}, \mathbf{T}, \mathbf{M}).
\end{equation}

To address the challenges posed by expensive evaluation and inefficient exploration in large agent design spaces, we propose a unified search framework that integrates a predictive \emph{value model} with an \emph{uncertainty-guided hierarchical expansion strategy} based on MCTS. The value model serves as a surrogate evaluator, estimating the performance of candidate agents based on their architecture and task description. This significantly reduces reliance on costly real-world evaluations by allowing the search to be guided by low-cost predictions. To cope with the vast combinatorial search space defined by agentic workflows and functional components, we propose a hierarchical expansion that operates over two levels of abstraction: agentic workflow and functional components. The expansion process includes three operations-\emph{recombination}, \emph{mutation}, and \emph{refinement}—each applied to both levels. Crucially, we incorporate uncertainty estimation from the value model to prioritize exploration of regions where performance predictions are both high and uncertain. Together, the predictive modeling and uncertainty-aware MCTS enable scalable, sample-efficient discovery of high-performing LLM agents. The algorithm is presented in Appendix.

\subsection{Value model}
To efficiently guide the agent search process and reduce the reliance on expensive real-world evaluations, we propose a predictive value model that estimates the performance of a candidate agent $\mathbf{A} = (\mathbf{W}, \mathbf{M}, \mathbf{T}, \mathbf{P})$ on a given task $d$. The model is trained to approximate the evaluation function $\text{Eval}_d(\cdot)$ via supervised learning:
\begin{equation}
    \hat{v} = f_\theta(\mathbf{A}, d),
\end{equation}
where $f_\theta$ denotes the learned value model and $\hat{v}$ is the predicted performance score.

Prior works have leveraged powerful LLM like GPT-4o as in-context predictors for this task, where historical agent performance data is fed directly into the prompt to estimate the success of a new agent~\cite{shang2024agentsquare}. However, such in-context evaluation requires repeated invocation of large models during search, resulting in high computational overhead. In contrast, our approach distills this predictive capability into a lightweight, task-generalized value model, enabling fast and scalable inference with significantly lower cost.

\paragraph{Dataset construction.}
To construct a high-quality training dataset $\mathcal{D} = \{ (\mathbf{A}_i, d_i, v_i) \}_{i=1}^{N}$, we employ a two-stage process designed to ensure both broad coverage and discriminative diversity(Figure \ref{figure:data}):

\begin{enumerate}[leftmargin=*]
    \item \textbf{$t$-way combinatorial coverage}: We begin by generating an initial dataset using a $t=2$ covering array to exhaustively sample combinations of pairwise interactions among the four key elements of the agent design: $\mathbf{W}$, $\mathbf{M}$, $\mathbf{T}$, and $\mathbf{P}$. This ensures that all pairwise component interactions are represented at least once, promoting coverage.
    \item \textbf{Balanced Bayesian sampling}: We augment the initial dataset using a Balanced-Extreme Bayesian Optimization strategy. We fit a Gaussian Process (GP) surrogate over the discrete agent space, using a Hamming kernel. The posterior mean $\mu(\mathbf{A})$ and standard deviation $\sigma(\mathbf{A})$ are used to define two acquisition functions:
\begin{equation}
\begin{split}
    a_{\text{UCB}}(\mathbf{A}) &= \mu(\mathbf{A}) + \kappa \cdot \sigma(\mathbf{A}), \\
    a_{\text{LCB}}(\mathbf{A}) &= -\mu(\mathbf{A}) + \kappa \cdot \sigma(\mathbf{A}).
\end{split}
\end{equation}
    where $\kappa$ is an exploration coefficient. In each sampling round, we select a batch of $q$ new agent designs from the candidate pool $\mathcal{S}$:
    \begin{equation}
    q_{\text{high}} = \left\lceil \tfrac{q}{2} \right\rceil, \quad q_{\text{low}} = q - q_{\text{high}},
    \end{equation}
    where $q_{\text{high}}$ maximizes $a_{\text{UCB}}$ to explore high-performing regions and $q_{\text{low}}$ maximizes $a_{\text{LCB}}$ to explore potentially underperforming yet uncertain configurations. This dual exploration yields a diverse and discriminative dataset. We repeat this process until a total of 220 labeled samples are obtained, which are then randomly split into training, validation, and test sets with a ratio of 8:1:1.

\end{enumerate}

\paragraph{Model architecture and training.} 
We implement the value model using a pre-trained 7B language model augmented with lightweight adapter modules, enabling robust generalization across diverse tasks. The entire model is fine-tuned end-to-end on the constructed dataset using mean squared error (MSE) loss. 


\setlength{\tabcolsep}{1.2mm} 
\begin{table*}[t]
  \label{tab:benchmark_comparison}
  \centering
  \small
  {\caption{Performance comparison of our method against hand-crafted agents and agent search methods across seven diverse benchmarks using GPT-4o-mini. The results are averaged over three independent runs. Our method consistently achieves the best performance across all benchmarks.}
    \begin{tabular}{>{\raggedright\arraybackslash}llcccccccc}
    \toprule
    \multirow{2}{*}{\textbf{Baseline Type}} & \multirow{2}{*}{\textbf{Method}} & \multicolumn{2}{c}{\textbf{Embodied}} & \textbf{Math} & \textbf{Web} & \multicolumn{2}{c}{\textbf{Tool}} & \textbf{Game} \\
    \cmidrule(lr){3-9}
    & & \textbf{Alfworld} & \textbf{SciWorld} & \textbf{MATH} & \textbf{WebShop} & \textbf{M3Tool} & \textbf{Travel} & \textbf{PDDL} \\
    \midrule
    \multirow{9}{*}{\makecell{Hand-crafted\\Agents}} 
        & COT & 0.512$\pm$0.009 & 0.398$\pm$0.005 & 0.532$\pm$0.004 & 0.490$\pm$0.011 & 0.427$\pm$0.008 & 0.433$\pm$0.003 & 0.427$\pm$0.011 \\
        & CoTSC & 0.545$\pm$0.006 & 0.412$\pm$0.004 & 0.543$\pm$0.002 & 0.488$\pm$0.006 & 0.451$\pm$0.012 & 0.410$\pm$0.001 & 0.410$\pm$0.009 \\
        & TOT & 0.530$\pm$0.008 & 0.384$\pm$0.004 & 0.547$\pm$0.005 & 0.462$\pm$0.009 & 0.463$\pm$0.014 & 0.407$\pm$0.007 & 0.433$\pm$0.007 \\
        & FoA & 0.587$\pm$0.005 & 0.427$\pm$0.008 & 0.556$\pm$0.003 & 0.509$\pm$0.012 & 0.488$\pm$0.009 & 0.474$\pm$0.006 & 0.472$\pm$0.007 \\
        & TP & 0.373$\pm$0.010 & 0.195$\pm$0.009 & 0.543$\pm$0.001 & 0.343$\pm$0.013 & 0.402$\pm$0.007 & 0.387$\pm$0.008 & 0.440$\pm$0.005 \\
        & SelfRefine & 0.575$\pm$0.007 & 0.375$\pm$0.006 & 0.551$\pm$0.004 & 0.425$\pm$0.010 & 0.463$\pm$0.010 & 0.047$\pm$0.015 & 0.412$\pm$0.008 \\
        & Dilu & 0.451$\pm$0.009 & 0.358$\pm$0.008 & 0.545$\pm$0.003 & 0.492$\pm$0.008 & 0.476$\pm$0.011 & 0.360$\pm$0.009 & 0.417$\pm$0.006 \\
        & Voyager & 0.336$\pm$0.011 & 0.389$\pm$0.005 & 0.517$\pm$0.006 & 0.423$\pm$0.012 & 0.317$\pm$0.014 & 0.517$\pm$0.004 & 0.337$\pm$0.010 \\
        & DEPS & 0.493$\pm$0.007 & 0.435$\pm$0.007 & 0.513$\pm$0.005 & 0.308$\pm$0.015 & 0.329$\pm$0.013 & 0.523$\pm$0.003 & 0.463$\pm$0.007 \\
        & Stepback & 0.470$\pm$0.008 & 0.314$\pm$0.009 & 0.530$\pm$0.002 & 0.459$\pm$0.011 & 0.488$\pm$0.009 & 0.033$\pm$0.012 & 0.403$\pm$0.009 \\
    \midrule
    \multirow{4}{*}{Agent Search} 
        & AgentSquare & \underline{0.701$\pm$0.07} & \underline{0.475$\pm$0.005} & 0.556$\pm$0.004 & \underline{0.520$\pm$0.009} & \underline{0.561$\pm$0.010} & \underline{0.553$\pm$0.004} & \underline{0.577$\pm$0.008} \\
        & AFlow & 0.619$\pm$0.006 & 0.452$\pm$0.007 & 0.562$\pm$0.003 & 0.497$\pm$0.011 & 0.524$\pm$0.012 & 0.497$\pm$0.006 & 0.528$\pm$0.008 \\
        & ADAS & 0.567$\pm$0.009 & 0.463$\pm$0.006 & 0.543$\pm$0.005 & 0.436$\pm$0.013 & 0.500$\pm$0.011 & 0.453$\pm$0.007 & 0.509$\pm$0.009 \\
        & MaAS & 0.612$\pm$0.007 & 0.437$\pm$0.008 & \underline{0.597$\pm$0.001} & 0.485$\pm$0.010 & 0.537$\pm$0.010 & 0.403$\pm$0.005 & 0.564$\pm$0.007 \\
    \midrule
    & \textbf{AgentSwift} & \textbf{0.806$\pm$0.007} & \textbf{0.509$\pm$0.006} & \textbf{0.628$\pm$0.000} & \textbf{0.562$\pm$0.010} & \textbf{0.634$\pm$0.013} & \textbf{0.573$\pm$0.002} & \textbf{0.614$\pm$0.008} \\
    \bottomrule
  \end{tabular}
  \label{tab:main}}
\end{table*}

\subsection{Uncertainty-guided MCTS}

\paragraph{Initialization.} 
To warm-start the search and improve early-stage efficiency, we initialize a global experience pool $\mathbb{E} = \{(\mathbf{W}, \mathbf{M}, \mathbf{T}, \mathbf{P}, v)\}$, where $v$ is the measured performance of an agent. This pool is seeded using well-designed baseline agents adapted from the AgentSquare~\cite{shang2024agentsquare} codebase. The pools $\{\mathbb{W}, \mathbb{M}, \mathbb{T}, \mathbb{P}\}$ are extracted from these baselines and standardized. 

\paragraph{Selection.}
We adopt a soft mixed probability selection strategy that integrates observed performance and model uncertainty, encouraging balanced exploration and exploitation. Given a set of $n$ candidate agents, the selection probability for agent $i$ is computed as:
\begin{equation}
E(s_j, u_j) = \alpha \cdot \left((1 - \beta) \cdot s_j + \beta \cdot u_j - s_{\text{max}}\right),
\end{equation}
\begin{equation}
P_{\text{mixed}}(i) = \lambda \cdot \frac{1}{n} + (1 - \lambda) \cdot \frac{\exp\left(E(s_i, u_i)\right)}{\sum_{j=1}^{n} \exp\left(E(s_j, u_j)\right)}.
\end{equation}
where $s_i$ denotes the actual task performance of agent $i$, $u_i$ is the uncertainty, and $s_{\text{max}}$ is the maximum composite score across all candidates. $\lambda$, $\alpha$, and $\beta$ control the trade-off between uniform exploration, sensitivity to performance differences, and the contribution of uncertainty, respectively. This formulation extends AFlow’s~\cite{zhang2024aflow} approach by incorporating epistemic uncertainty, thus encouraging the search to explore candidates that are either high-performing or insufficiently evaluated.

\paragraph{Expansion.}
Starting from the parent agent returned by selection, we perform a top-down \emph{hierarchical expansion} composed of three operations—\emph{recombination}, \emph{mutation}, and \emph{refinement}. These operations utilize task-agnostic prompts, where only the task description varies for new applications. Each operation generates a small batch of new agents and the value model scores every candidate and the best one advances to the next operation.  
\begin{enumerate}[leftmargin=*,nosep]

\item \textbf{Recombination}\;
      An LLM proposer $\pi_{\theta}$ (adapted from AgentSquare) replaces one subsystem—\emph{agentic workflow}, \emph{planning}, \emph{tool use}, or \emph{memory}—with an alternative sampled from the corresponding pool.  
      Given a current agent $(\mathbf{W}_0,\mathbf{M}_0,\mathbf{T}_0,\mathbf{P}_0)$ and experience pool $\mathbb{E}$, $\pi_{\theta}$ produces $N$ candidate agents.  
      For example, $(\mathbf{W}_0,\mathbf{M}_0,\mathbf{T}^{'},\mathbf{P}_0)$ denotes a recombination where the tool use component is replaced with a new $\mathbf{T}^{'} \in \mathbb{T}$.
      The value model ranks all candidates, and the top one is passed to the next phase.


\item \textbf{Mutation}\;
      An LLM programmer $\pi_{\xi}$ generates a new implementation of the selected subsystem by leveraging task description, existing subsystems, and prior agent performance from $\mathbb{E}$.  
      This yields $N$ mutated agents; for instance, $(\mathbf{W}_0, \mathbf{M}_0, \mathbf{T}_0, \mathbf{P}^*)$ represents a mutated variant where a new planning $\mathbf{P}^*$ is synthesized.       The value model ranks all candidates, and the top one is passed to next phase.
      Newly generated subsystems are appended to the global pools so future searches can reuse them.


\item \textbf{Refinement}\;
      An LLM refiner $\pi_{\phi}$ applies fine-grained adjustments to the selected agent by modifying a single subsystem in light of failure cases.  
      These refinements include prompt edits, temperature nudges or control-flow modifications.  
      For example, $(\mathbf{W}^`, \mathbf{M}_0, \mathbf{T}_0, \mathbf{P}_0)$ denotes a refined variant with an updated workflow $\mathbf{W}^`$.  
      Among the refined candidates, the one with the highest predicted performance is inserted into the MCTS tree.

\end{enumerate}
This three-step pipeline simultaneously broadens exploration (via recombination), unlocks novel behaviour (via mutation), and polishes promising designs (via refinement)

\paragraph{Evaluation.}
Inspired by classic works on probabilistic forecasting and sequential decision-making \cite{brier1950verification, auer2002finite, kocsis2006bandit}, the child agent is evaluated on the target task to obtain its actual performance score $s_{\text{real}}$. To quantify the epistemic uncertainty of the value model's prediction, we define the uncertainty as the absolute deviation between the predicted score $\hat{s}$ and the true performance:
\begin{equation}
u = \bigl|s_{\text{real}} - \hat{s}\bigr|.
\end{equation}
This uncertainty metric, rooted in the principles of forecast calibration \cite{brier1950verification}, enables the search algorithm to balance exploitation of high-performing configurations with exploration of under-evaluated regions.

\paragraph{Backpropagation.} 
After evaluation, the node records its actual score $s_{\mathrm{real}}$ together with the uncertainty $u$.  
These statistics are then propagated upward, where each ancestor node increments its visit count.  
Finally, node $\bigl(\mathbf{W},\mathbf{M},\mathbf{T},\mathbf{P},s_{\text{real}}\bigr)$ is attached to the global experience pool~$\mathbb{E}$, enlarging the candidate set for subsequent iterations.

\section{Experiments}

\subsection{Experimental setup}

\paragraph{Task setup.}
We evaluate our framework on seven benchmark spanning five representative task domains commonly used in LLM evaluation~\cite{ma2024agentboard,xi2024agentgym}. More details are presented in Appendix.

\paragraph{Baselines.}
We compare our framework against two categories of baselines including manually designed agent and automated agent search methods. More details are presented in Appendix.

\paragraph{Implementation details.}
\label{Implementation}

We conduct experiments using closed-source LLMs (\texttt{gpt-4o~\cite{gpt4o2024openai}}, \texttt{gpt-4o-mini~\cite{achiam2023gpt}}) and an open-source one (\texttt{DeepSeek-v3~\cite{liu2024deepseek}}). For the value model, we adopt two backbone LLMs: \texttt{Mistral-7B-v0.3~\cite{jiang2024mistral}} and \texttt{Qwen2.5-7B~\cite{yang2024qwen2}}. To ensure fair comparison across agent search methods, the evaluation budget is capped at 60 agents per method, by which point all baselines converge. The value model is trained on a server equipped with 3 A100 GPUs.
\subsection{Experimental results}

\paragraph{Main results.}

\begin{figure}[t]
\begin{center}
\captionsetup{belowskip=1pt}
\includegraphics[width=\columnwidth]{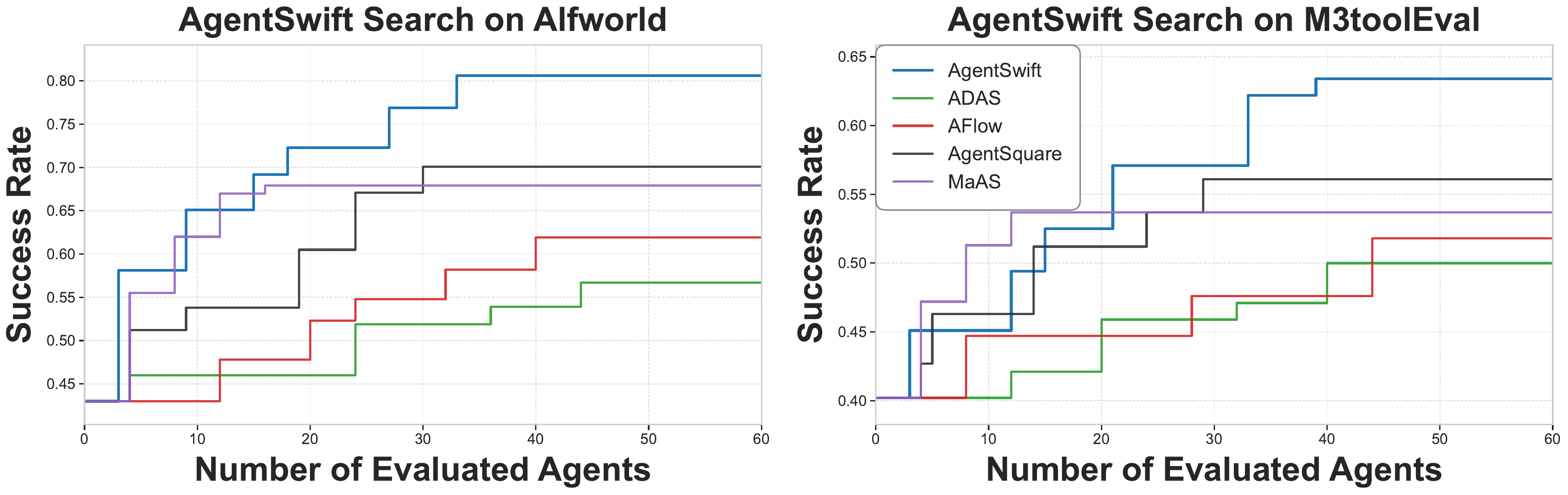}
\caption{AgentSwift search trajectory on Alfworld and M3ToolEval.}
\label{fig:trajectory_agent1}
\end{center}
\end{figure}

\begin{itemize}[leftmargin=*]
    \item \textbf{Our method consistently discovers the best-performing agents.} 
    Across all tasks, our framework reliably identifies agent designs that outperform both manually constructed baselines and agents discovered by existing search methods. As shown in Table~\ref{tab:main}, our best-found agents achieve substantial improvements over the strongest competing methods. These consistent gains highlight the advantage of searching over both agentic workflows and composable functional components in a unified hierarchical design space. Our formulation enables richer architectural compositions beyond fixed or manually selected modules, yielding more generalizable and effective agent behaviors.

    \item \textbf{Our method enables steeper and more efficient search trajectories.}   Figures~\ref{fig:trajectory_agent1} present the search trajectories of our method and strong baselines. Our method demonstrates a noticeably steeper and more stable performance curve across tasks, indicating faster discovery of high-performing agent. In contrast, methods such as AFlow and ADAS either stagnate due to limited agentic workflow variation or require significantly more iterations and time to escape local optima. These results validate the synergistic effectiveness of predictive modeling and structured, uncertainty-aware search in accelerating agent discovery. For brevity, search trajectories for other tasks, alongside detailed analyses of wall-clock time and the cost-performance Pareto front, are provided in the Appendix.


\end{itemize}




\begin{figure}[t]
\begin{center}
\includegraphics[width=\columnwidth]{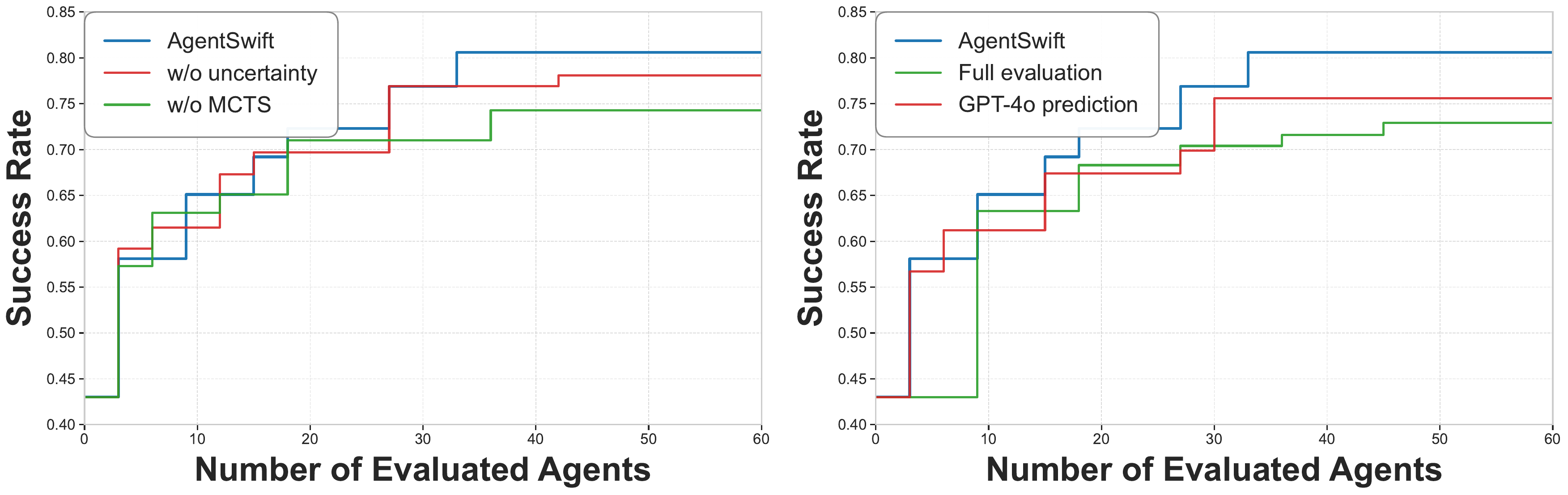}
\captionsetup{belowskip=3pt}
\caption{
Left: search trajectory of different search strategies on Alfworld: AgentSwift, w/o uncertainty, and w/o MCTS. Right: search trajectory of different evaluate method on Alfworld: AgentSwift, gpt-4o prediction, and full evaluation.
}
\label{fig:search strategies and predictor}
\end{center}
\end{figure}

\begin{table}[t]
  \captionsetup{belowskip=5pt}
  \caption{Performance comparison of different surrogate models on all benchmarks. Our method consistently achieves the best performance across all metrics.}
  \label{tab:value model}
  \centering
  \small
  \begin{tabular}{lcccc}
    \toprule
    \textbf{Method} & \textbf{MSE} & \textbf{MAE} & \textbf{R²} & \textbf{Spearman} \\
    \midrule
    \textbf{AgentSwift$_{\text{mistral}}$} & \textbf{0.0060} & \textbf{0.0530} & \textbf{0.8068} & \textbf{0.9026} \\
    \textbf{AgentSwift$_{\text{qwen}}$} & \textbf{0.0054} & \textbf{0.0547} & \textbf{0.8275} & \textbf{0.8987} \\
    vanilla & 0.1572 & 0.3593 & -4.0590 & 0.2467 \\
    gpt-4o few shot & 0.0162 & 0.0893 & 0.4793 & 0.7654 \\
    gpt-4o zero shot & 0.0675 & 0.2067 & -1.1708 & 0.0563 \\
    gpt-4o-mini few shot & 0.0307 & 0.1179 & 0.0114 & 0.5410 \\
    gpt-4o-mini zero shot & 0.0820 & 0.2370 & -1.6403 & -0.0774 \\
    \bottomrule
  \end{tabular}
\end{table}

\paragraph{Analysis of value model.}
We evaluate the effectiveness of our value model by comparing it with several baseline predictors trained on the same dataset, including a vanilla supervised model and in-context learning methods using GPT-4o and GPT-4o-mini in both zero-shot and few-shot settings. As shown in Table ~\ref{tab:value model}, our approach achieves the best performance across all metrics—MSE, MAE, $R^2$, and Spearman correlation—demonstrating superior accuracy in both absolute prediction and ranking quality. 


\paragraph{Analysis of search strategy.}
We analyze the effect of our search design by comparing variants of our method on the AlfWorld. As shown in Figure ~\ref{fig:search strategies and predictor} (Left), removing MCTS or uncertainty guidance significantly flattens the search trajectory. Without MCTS, the algorithm lacks hierarchical exploration and becomes overly local, while without uncertainty, the search tends to exploit familiar regions and misses promising but uncertain candidates. On the right, we compare different evaluation strategies. Our value model enables faster improvement than GPT-4o few-shot due to its higher prediction accuracy. In contrast, full evaluation progresses the slowest, as it exhausts much of the evaluation budget on low-performing agents. These results highlight the importance of accurate value estimation and selective evaluation in enabling efficient and targeted agent discovery.



\paragraph{Model-agnostic.} 
To assess the transferability of discovered agents across different LLMs, we perform agent search using \texttt{gpt-4o-mini} and then directly evaluate the resulting agent architectures on other models. Our framework demonstrates strong cross-model transferability, as detailed in the table presented in the Appendix.

\paragraph{Hyperparameter Sensitivity.}
We analyze the sensitivity of our search strategy's key hyperparameters: $\alpha$, $\lambda$, and $\beta$. Our default configuration uses $\alpha=3.0$, $\lambda=0.3$, and $\beta=0.4$. As shown in Table~\ref{tab:hyperparameter}, we varied each parameter individually while holding the others constant, evaluating performance on the Alfworld benchmarks. The results demonstrate that AgentSwift is robust to variations in these hyperparameters, maintaining strong performance across a range of values.

\begin{table}[t]
 \centering
 \small
 \captionsetup{belowskip=3pt}
 \caption{Sensitivity analysis of hyperparameters $\alpha$, $\lambda$, and $\beta$ on Alfworld benchmarks. Performance is robust across different settings.}
 \label{tab:hyperparameter}
  \begin{tabular}{lcc}
 \toprule
   \textbf{Hyperparameter} & \textbf{Alfworld} \\
\midrule
 $\alpha=2.0$ & 0.784  \\
 $\alpha=3.0$ (default) & 0.806 \\
 $\alpha=4.0$ & 0.813  \\
 $\alpha=5.0$ & 0.799  \\
 \midrule
 $\lambda=0.1$ & 0.768  \\
 $\lambda=0.2$ & 0.795  \\
 $\lambda=0.3$ (default) & 0.806 \\
 $\lambda=0.4$ & 0.784 \\
 \midrule
 $\beta=0.2$ & 0.793  \\
 $\beta=0.3$ & 0.785  \\
 $\beta=0.4$ (default) & 0.806  \\
 $\beta=0.5$ & 0.801 \\
 \bottomrule
 \end{tabular}
\end{table}

\begin{figure}[!t]
\begin{center}
\captionsetup{belowskip=3pt}
\includegraphics[width=0.4\textwidth]{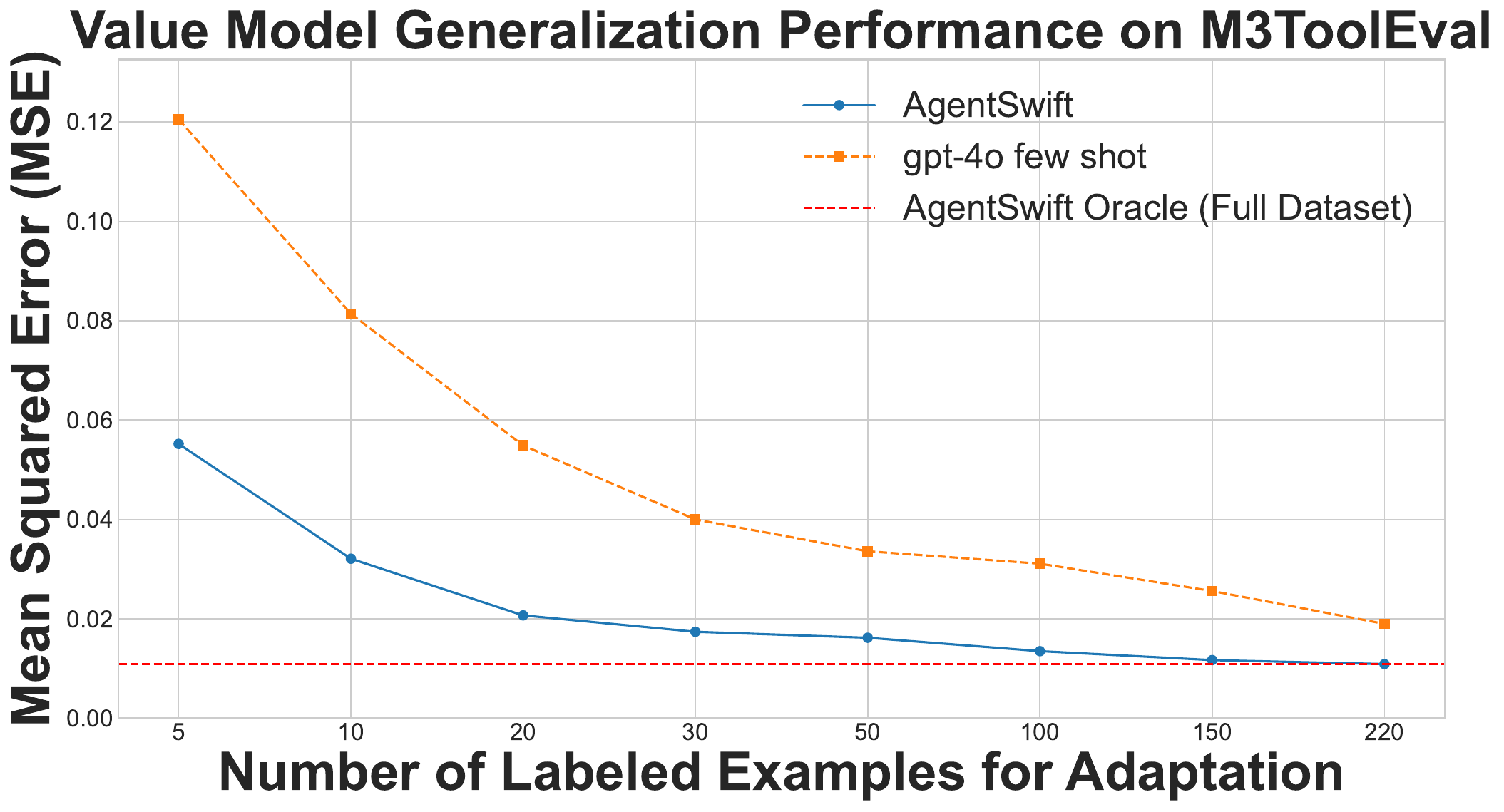}
\caption{Performance comparison on M3ToolEval under few-shot adaptation.}
\label{fig:generalization}
\end{center}
\end{figure}


\subsection{Generalization analysis}
We evaluate the generalization ability of our value model by adapting it to the unseen M3ToolEval benchmark using a varying number of labeled examples for few-shot adaptation. As shown in Figure~\ref{fig:generalization}, which plots MSE against the number of adaptation samples, our value model demonstrates remarkable sample efficiency. With as few as 30 labeled examples, our model's performance already approaches the oracle performance achieved when trained on the full dataset. This strong generalization capability is attributed to the highly structured agent representation: the hierarchical design of agentic workflow and functional components forms a compositional abstraction that is both interpretable and transferable across tasks. This allows the model to learn the relationship between agent and performance effectively, even with minimal supervision on a new task.

\subsection{Ablation study}
To assess the contribution of each stage in our hierarchical search strategy, we conduct ablations by individually removing the recombination, mutation, and refinement stages. More details are presented in Appendix.

\section{Conclusion}
In this work, we propose a unified framework for automated agentic system search that combines a hierarchical search space with a value model and an uncertainty-guided hierarchical MCTS strategy. Our formulation captures both the structural workflow and functional components of agents, enabling rich architectural variation and compositional reasoning. The value model provides accurate and low-cost performance prediction, while the uncertainty-aware search strategy efficiently explores the vast design space by prioritizing promising candidates.

\bibliography{aaai2026}

\onecolumn  
\appendix
\section{Appendix}
\renewcommand{\thefigure}{A.\arabic{figure}}
\renewcommand{\thetable}{A.\arabic{table}}
\subsection{Experimental setup}
\label{app:setup}
\paragraph{Task setup.}
We evaluate our framework on seven benchmark spanning five representative task domains commonly used in LLM agent evaluation~\cite{ma2024agentboard,xi2024agentgym}. These include (1) \textbf{Embodied}: ALFWorld~\cite{shridhar2021alfworld} and ScienceWorld~\cite{wang2022scienceworld}, which require agents to navigate and manipulate environments using text commands; (2) \textbf{Game}: PDDL~\cite{ma2024agentboard}, which includes many strategic games where agents use PDDL expressions to complete task; (3) \textbf{Web}: WebShop~\cite{yao2022webshop}, where agents perform goal-directed browsing and purchasing; (4) \textbf{Tool}: TravelPlanner~\cite{xie2024travelplanner} and M3ToolEval~\cite{wang2024executable}, which test multi-tool integration; and (5) \textbf{Math}: MATH~\cite{hendrycks2024measuring}, a benchmark of grade-school to olympiad-level problems requiring step-wise symbolic reasoning. For the MATH dataset, we adopt the subset construction strategy proposed by Hong et al.~\cite{hong2024data}, selecting 617 problems from four representative categories—Combinatorics \& Probability, Number Theory, Pre-algebra, and Pre-calculus—all at difficulty level 5.


\paragraph{Baseline.}
We compare our framework against two categories of baselines: manually designed agent systems that represent popular, fixed strategies, and recent automated agent search methods that explore the design space.

\begin{itemize}
    \item \textbf{Manually Designed Agent Systems:} These methods employ predefined reasoning structures and workflows.
    \begin{itemize}
        \item \textbf{Chain-of-Thought (CoT)}~\cite{wei2022chain} prompts the LLM to generate a sequence of intermediate reasoning steps to solve a problem, rather than producing an answer directly.
        \item \textbf{Self-Consistency (CoTSC)}~\cite{wang2022self} improves upon CoT by sampling multiple diverse reasoning paths and selecting the most consistent answer through a majority vote.
        \item \textbf{Tree-of-Thought (ToT)}~\cite{yao2023tree} generalizes CoT by organizing thoughts in a tree structure, allowing the model to explore, evaluate, and backtrack among different reasoning branches.
        \item \textbf{Fleet-of-Agents (FoA)}~\cite{klein2024fleet} proposes an evolutionary method to create a diverse "fleet" of agents, where different agent workflows are generated and then composed to collaboratively solve a problem.
        \item \textbf{Thought Propagation}~\cite{yu2024thought} introduces an analogical reasoning framework where the model leverages insights from related solved problems to guide its process on new, complex tasks.
        \item \textbf{Self-Refine}~\cite{madaan2023self} uses an iterative feedback loop where the same LLM refines its own previously generated outputs, correcting mistakes and improving quality without external supervision.
        \item \textbf{DILU}~\cite{wen2024dilu} is a knowledge-driven agent designed for autonomous driving.
        \item \textbf{Voyager}~\cite{wang2024voyager} is an embodied agent for open-ended exploration in Minecraft, which autonomously builds a skill library.
        \item \textbf{DEPS}~\cite{wang2024describe} is an interactive planning framework that enables agents to handle complex, open-world tasks through a cycle of Describing, Explaining, Planning, and Selecting actions.
        \item \textbf{Step-Back Planning}~\cite{zheng2024take} enhances reasoning by prompting an LLM to first abstract away from specific details to derive high-level concepts, which then guide the detailed reasoning steps.
    \end{itemize}
    \item \textbf{Automated Agent Search Methods:} These frameworks aim to automatically discover effective agent architectures or workflows.
    \begin{itemize}
        \item \textbf{AgentSquare}~\cite{shang2024agentsquare} automates agent design by searching within a modular space under a fixed workflow template.
        \item \textbf{AFlow}~\cite{zhang2024aflow} formulates agentic workflow generation as a search problem over a graph, where nodes represent LLM calls and edges define the execution flow.
        \item \textbf{ADAS}~\cite{hu2024automated} proposes an algorithm to automatically search for and design effective agentic systems.
        \item \textbf{MaAS}~\cite{zhang2025multi} shifts from searching for a single optimal architecture to learning a policy that generates a query-conditioned distribution over different multi-agent system designs.
    \end{itemize}
\end{itemize}

\begin{algorithm}[H]
\caption{Algorithm of AgentSwift}
\label{alg:agentsearch}
\KwIn{Task $d$, budget $B$, seed set $\mathbb{E}_0$, module pools $\{\mathbb{W}, \mathbb{M}, \mathbb{T}, \mathbb{P}\}$}
\KwOut{Best agent $\mathbf{A}^*$}

Initialize experience pool $\mathbb{E} \leftarrow \mathbb{E}_0$, train value model $f_\theta$ on $\mathbb{E}$\;
Initialize $\mathbf{A}^* \leftarrow \arg\max_{\mathbf{A} \in \mathbb{E}} \text{Eval}_d(\mathbf{A})$\;

\For{$t = 1$ \KwTo $B$}{
    \tcp{Selection}
    Sample $\mathbf{A}_p = (\mathbf{W}, \mathbf{M}, \mathbf{T}, \mathbf{P}) \sim P_{\text{mixed}}$ over $\mathbb{E}$ where:
    \[
    P_{\text{mixed}}(i) = \lambda \cdot \tfrac{1}{n} + (1 - \lambda) \cdot \frac{\exp\left(\alpha ((1 - \beta) s_i + \beta u_i - s_{\max})\right)}{\sum_j \exp\left(\alpha ((1 - \beta) s_j + \beta u_j - s_{\max})\right)}
    \]

    \tcp{Expansion: Recombination}
    $\{\mathbf{A}_{\text{rec}}^{(k)}\}_{k=1}^N \leftarrow \pi_\theta(\mathbf{A}_p, \{\mathbb{W}, \mathbb{M}, \mathbb{T}, \mathbb{P}\}, \mathbb{E})$\;
    $\mathbf{A}_{\text{rec}} \leftarrow \arg\max_k f_\theta(\mathbf{A}_{\text{rec}}^{(k)}, d)$

    \tcp{Expansion: Mutation}
    $\{\mathbf{A}_{\text{mut}}^{(k)}\}_{k=1}^N \leftarrow \pi_\xi(\mathbf{A}_{\text{rec}}, d, \mathbb{E})$\;
    $\mathbf{A}_{\text{mut}} \leftarrow \arg\max_k f_\theta(\mathbf{A}_{\text{mut}}^{(k)}, d)$\;
    Add new modules to global pools

    \tcp{Expansion: Refinement}
    $\{\mathbf{A}_{\text{ref}}^{(k)}\}_{k=1}^N \leftarrow \pi_\phi(\mathbf{A}_{\text{mut}}, \text{failures})$\;
    $\mathbf{A}_{\text{ref}} \leftarrow \arg\max_k f_\theta(\mathbf{A}_{\text{ref}}^{(k)}, d)$

    \tcp{Evaluation}
    Evaluate agent: $s_{\text{real}} \gets \text{Eval}_d(\mathbf{A}_{\text{ref}})$\;
    Predict: $\hat{s} \gets f_\theta(\mathbf{A}_{\text{ref}}, d)$\;
    Compute uncertainty: $u \gets |\hat{s} - s_{\text{real}}|$\;
    
    \tcp{Backpropagation}
    Update $\mathbb{E} \leftarrow \mathbb{E} \cup \{(\mathbf{A}_{\text{ref}}, s_{\text{real}}, u)\}$\;
    Update MCTS tree statistics\;
    $\mathbf{A}^* \leftarrow \arg\max\{s_{\text{real}}, \text{Eval}_d(\mathbf{A}^*)\}$
}
\Return{$\mathbf{A}^*$}
\end{algorithm}

\subsection{Additional Results on Search Efficiency}
\label{sec:appendix_efficiency}

As stated in the main paper, our framework demonstrates superior search efficiency across multiple dimensions. This section provides the detailed figures and analyses that were omitted for brevity.

\begin{figure}[H]
\begin{center}
\includegraphics[width=0.9\textwidth]{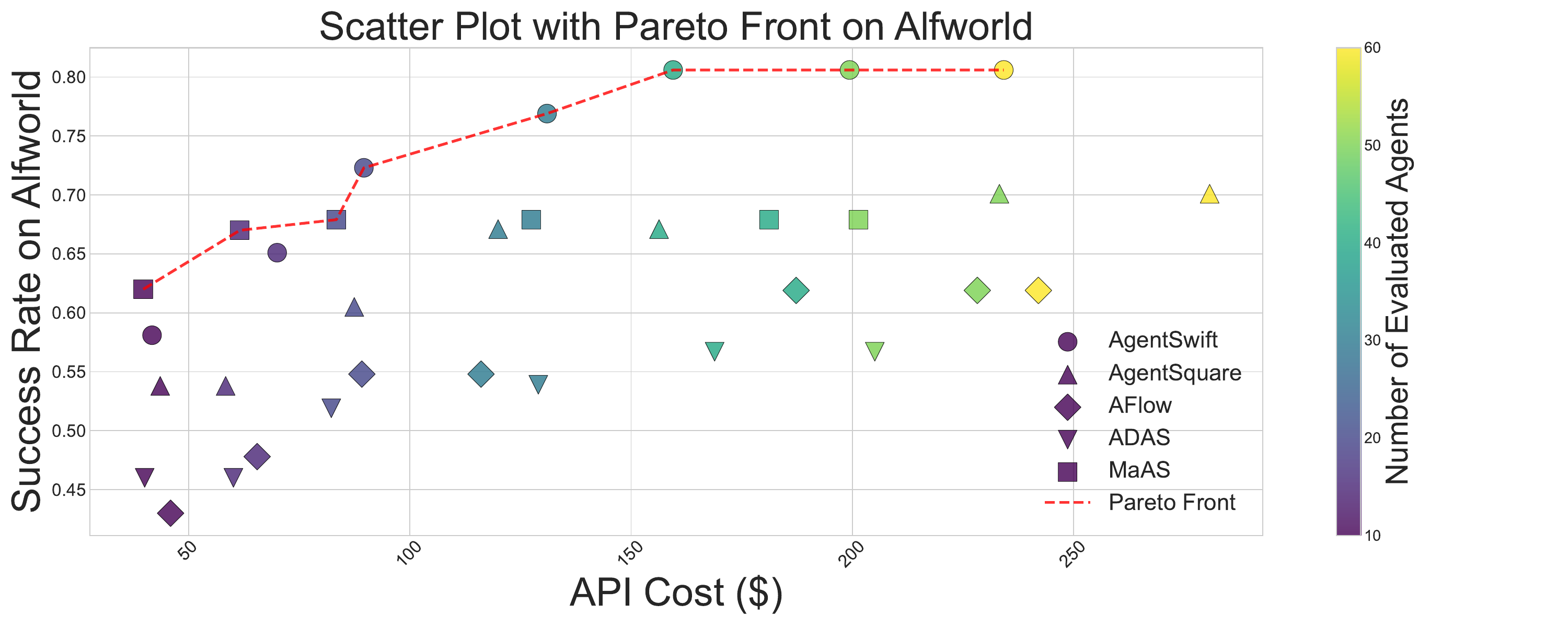}
\caption{Cost-Performance Pareto Front Analysis on Alfworld.}
\label{fig:pareto}
\end{center}
\end{figure}

\begin{figure}[H]
\begin{center}
\includegraphics[width=0.9\textwidth]{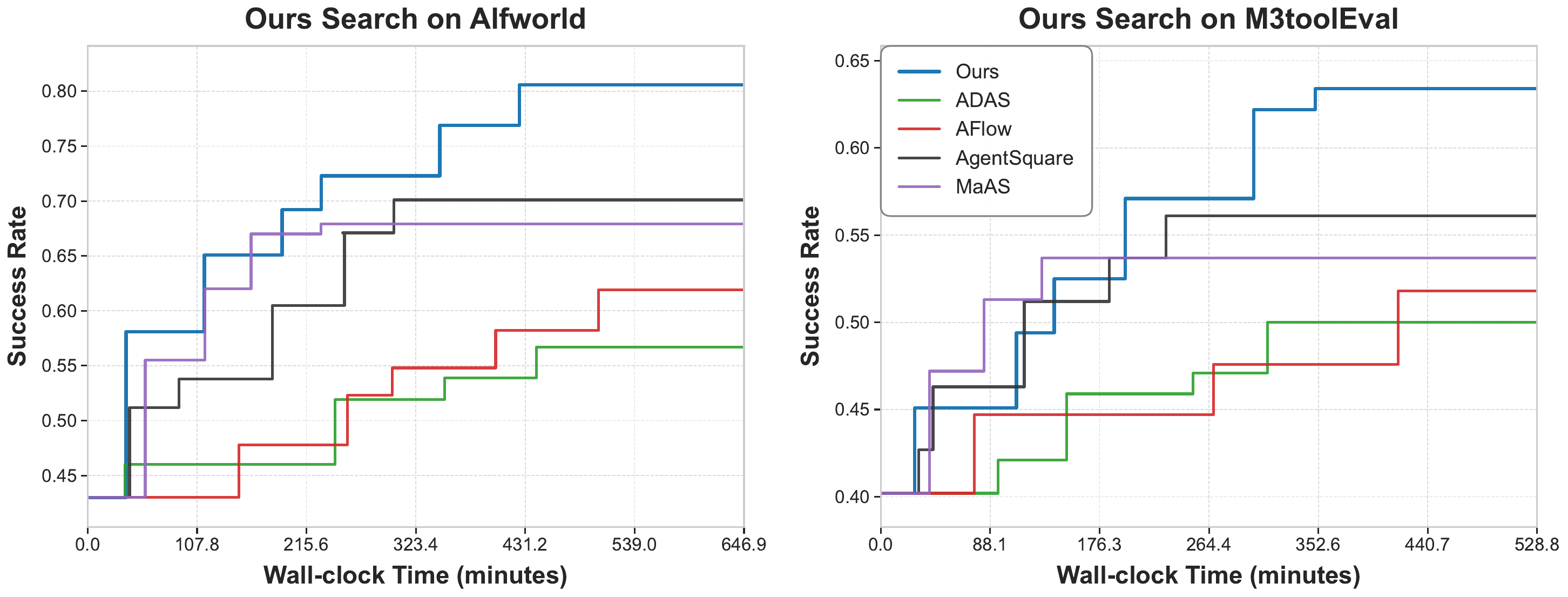}
\caption{AgentSwift search trajectory on Alfworld and M3ToolEval.}
\label{fig:trajectory_time1}
\end{center}
\end{figure}

\begin{figure}[H]
\begin{center}
\includegraphics[width=0.9\textwidth]{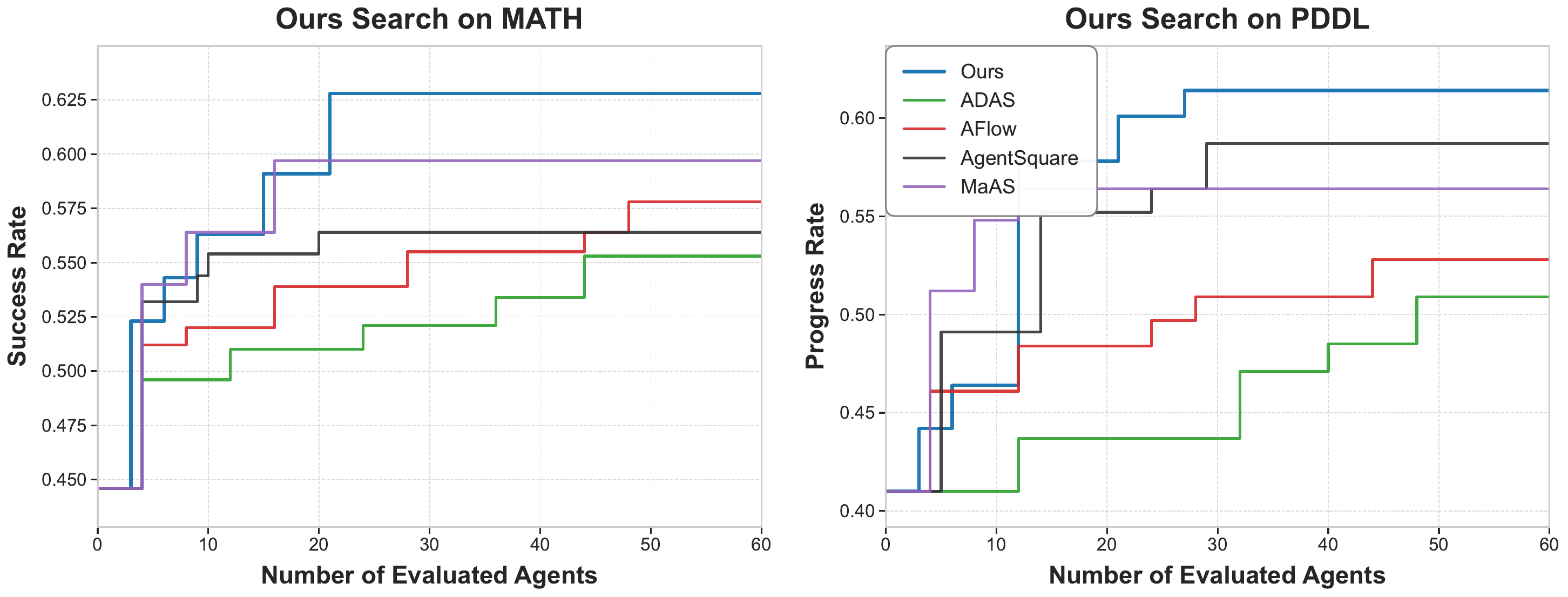}
\caption{AgentSwift search trajectory on MATH and PDDL.}
\label{fig:trajectory_agent2}
\end{center}
\end{figure}

\begin{figure}[H]
\begin{center}
\includegraphics[width=0.9\textwidth]{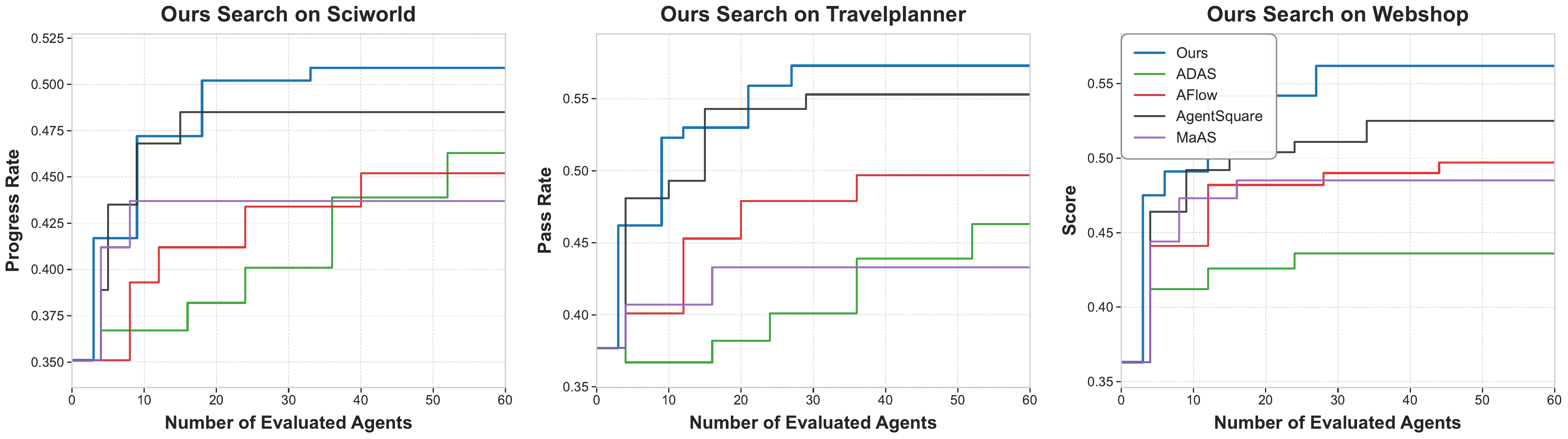}
\caption{AgentSwift search trajectory on Alfworld, Travelplanner and Webshop.}
\label{fig:trajectory_agent3}
\end{center}
\end{figure}

\begin{itemize}[leftmargin=*]
    \item \textbf{Search Trajectories on Additional Benchmarks.} Figures~\ref{fig:trajectory_agent2} and~\ref{fig:trajectory_agent3} present the search trajectories on the remaining five benchmarks: MATH, PDDL, Sciworld, Travelplanner, and Webshop. Consistent with the results on Alfworld and M3ToolEval, \texttt{AgentSwift} exhibits a steeper and more stable learning curve across all tasks. Our method consistently discovers agents with higher performance using significantly fewer evaluations than all baseline methods. This underscores the robustness and general applicability of our hierarchical search space and uncertainty-guided exploration strategy.
    \item \textbf{Wall-Clock Time Analysis.} To assess real-world efficiency, we analyze the performance improvement over wall-clock time. Figure~\ref{fig:trajectory_time1} plots the success rate against execution time (in minutes) for the Alfworld and M3ToolEval tasks. The results clearly show that \texttt{AgentSwift} not only reaches higher performance but does so in substantially less time.
    \item \textbf{Cost-Performance Pareto Front Analysis.} For the most direct analysis of computational cost, we present a Pareto front analysis in Figure~\ref{fig:pareto}. This plot maps agent performance (Success Rate) against the simulated cumulative API cost, providing a clear view of the cost-performance trade-off on the Alfworld benchmark. The Pareto front, highlighted by the red dashed line, identifies the set of agents offering the maximum performance for a given cost budget. The analysis yields a critical insight: while a baseline method like \texttt{MaAS} is optimal for scenarios with very strict, low-cost budgets, \textbf{our method, \texttt{AgentSwift}, exclusively constitutes the entire high-performance segment of the frontier}. This demonstrates that to achieve any state-of-the-art performance level (e.g., a success rate above 0.769), \texttt{AgentSwift} is the most resource-efficient—and often the only—viable option. The ability of our framework to dominate the high-performance region stems directly from the synergistic effect of our value-guided search, which efficiently prunes the vast design space and allocates computational resources toward discovering genuinely superior agent architectures. This validates the core claim that our method is not only effective but also highly cost-efficient.
\end{itemize}



\begin{table}[H]
  \centering
  \caption{Ablation study of our method}
    \label{tab:ablation}
  \begin{tabular}{lccc}
    \toprule
    \textbf{Method} & \textbf{ALFWorld} & \textbf{MATH} \\
    \midrule
    AgentSwift(full) & \textbf{0.806} & \textbf{0.628} \\
    w/o recombination & 0.739 & 0.588 \\
    w/o mutation & 0.761 & 0.562 \\
    w/o refinement & 0.776 & 0.621 \\
    \bottomrule
  \end{tabular}
  \label{tab:tab1}
\end{table}

\subsection{Ablation study}
To assess the contribution of each stage in our hierarchical search strategy, we conduct ablations by individually removing the recombination, mutation, and refinement stages. As shown in Table ~\ref{tab:ablation}, removing any single stage leads to a noticeable performance drop on both ALFWorld and MATH. The absence of recombination has the most pronounced effect, reducing performance from 0.806 to 0.739 on ALFWorld. Removing mutation also degrades results significantly, as it restricts the exploration of novel design variations beyond simple recombination. Excluding refinement results in a smaller decline, suggesting that feedback-driven adjustments help correct failure modes.

\clearpage
\begin{table}[H]
\vspace{0cm}
\caption{Cross-model transferability. We perform agent search using \texttt{gpt-4o-mini}, and evaluate the resulting agent by deploying them on other LLM.}
\label{tab:cross-llm}
\vspace{0.1em}
\centering
\begin{tabular}{c|ccc}
\toprule
Benchmark &\multicolumn{3}{c}{ALFworld}\\
\midrule
Method  &  gpt-4o-mini & DeepSeek-V3  & gpt-4o  \\
\midrule
Aflow    & $0.619$ & $0.746$ & $0.724$ \\
AgentSquare    & $0.701$ & $0.769$ & $0.761$ \\
MaAS    & $0.612$ & $0.709$ & $0.701$ \\
AgentSwift    & $0.806$ & $0.843$ & $0.851$ \\

\midrule

Benchmark &\multicolumn{3}{c}{MATH}\\
\midrule
Method  &  gpt-4o-mini & DeepSeek-V3  & gpt-4o \\
\midrule
Aflow    & $0.562$ & $0.881$ & $0.583$ \\
AgentSquare    & $0.556$ & $0.852$ & $0.574$ \\
MaAS    & $0.597$ & $0.874$ & $0.601$ \\
AgentSwift    & $0.628$ & $0.916$ & $0.630$ \\

\midrule

Benchmark &\multicolumn{3}{c}{M3tooleval}\\
\midrule
Method  &  gpt-4o-mini & DeepSeek-V3  & gpt-4o \\
\midrule
Aflow    & $0.524$ & $0.585$ & $0.609$ \\
AgentSquare    & $0.561$ & $0.609$ & $0.622$ \\
MaAS    & $0.537$ & $0.573$ & $0.549$ \\
AgentSwift    & $0.634$ & $0.695$ & $0.683$ \\
\bottomrule
\end{tabular}
\end{table}
\vspace{-0.5cm}

\begin{table}[H]
    \caption{Performance comparison on M3ToolEval under few-shot adaptation. Our model, trained on a different task and adapted to M3ToolEval using only 30 examples via task-specific adapters, consistently outperforms all baselines, demonstrating strong generalization and data efficiency.}
    \label{tab:generalization}
    \centering
    \begin{tabular}{lcccc}
      \toprule
      Method & MSE & MAE & R² & Spearman \\
      \midrule
      AgentSwift & 0.0174 & 0.0924 & 0.0526 & 0.4454 \\
      vanilla & 0.1744 & 0.3950 & -8.4970 & 0.2111 \\
      gpt-4o few shot & 0.0400 & 0.1592 & -1.1793 & 0.2738 \\
      gpt-4o zero shot & 0.0429 & 0.1741 & -1.3386 & 0.0656 \\
      gpt-4o-mini few shot & 0.0493 & 0.1382 & -0.5948 & 0.2504 \\
      gpt-4o-mini zero shot & 0.1070 & 0.2845 & -4.8279 & -0.2805 \\
      \bottomrule
    \end{tabular}
  \end{table}
  \vspace{-0.5cm}

\subsection{Limitations}
\label{Limitations}
While our framework demonstrates strong performance and efficiency in agentic system search, several limitations remain. First, the current mechanism for proposing new agents—through recombination, mutation, and refinement—is largely heuristic and relies on manually crafted LLM prompts. A more principled alternative would be to directly train or fine-tune a model that maps from a task specification (or query) to a compatible agentic system, effectively learning to generate well-adapted agents in a single forward pass. Although the value model generalizes across tasks with lightweight adapter tuning, it still requires labeled training data, which can be expensive to obtain in highly specialized or interactive environments. Incorporating self-supervised or preference-based signals could further reduce the need for real-task annotations. The current search framework focuses on static agent design and does not account for dynamic adaptation during deployment. Enabling agents to revise or recompose their workflows based on execution feedback in real time would be a promising direction for building more adaptive and robust systems.

\subsection{Broader impact}
\label{broader impact}
This work proposes an efficient framework for agentic system search, potentially benefiting applications in education, research, and productivity by reducing the cost of discovering high-performing LLM agents. By improving search efficiency, it enables broader access to advanced agent designs, especially for users with limited computational resources. However, the enhanced capability of agents may also introduce risks of misuse, such as disinformation or unauthorized data access. While our framework is task-agnostic and does not support harmful behavior, we encourage responsible use and the integration of appropriate safeguards. This work complies with the NeurIPS Code of Ethics and does not pose foreseeable societal harm when used appropriately.

\begin{tcolorbox}[colback=gray!10, colframe=black!50!black, title=The new agent found during the search on ALFWorld]
\lstset{basicstyle=\scriptsize\ttfamily} 
\begin{lstlisting}

class WorkflowSelfReflectiveTOT(WorkflowBase):
    def __call__(self, task_description: str, feedback :str= ''):
        examples, task_description = self.process_task_description(task_description)
        prompt = '''Your instructions must follow the examples.
Here are some examples.{examples}{memory}Here is the task:
{task_description}'''
        prompt = prompt.format(task_description=task_description, examples=examples, memory=self.memory_cache)
        workflow_results = llm_response(prompt=prompt, model=self.llm_type, temperature=0.1, stop_strs=['\n'], n=5)
        workflow_result = self.get_votes(task_description, workflow_results, examples)
        workflow_result = self.refine(Workflow_result)
        return Workflow_result
    def get_votes(self, task_description, workflow_results, examples):
        ...
    def refine(self, workflow_result):
        ...
class ConciseClarityPlanning(PlanningBase):
    def create_prompt(self, task_type, task_description, feedback, few_shot):
        prompt = '''You are a planner tasked with breaking down a {task_type} task into clear and concise subtasks. For each subtask, provide a straightforward description, reasoning instructions that are brief and closely follow the examples, and specify any necessary tools. Ensure your output is structured and mirrors the examples precisely.
Your output format should follow the example below.The following are some examples:Task{example} Task: {task_description}'''
        prompt = prompt.format(example=few_shot, task_description=task_description, task_type=task_type)
class ContextualGuidanceMemory(MemoryBase):
    def retriveMemory(self, query_scenario):
        # Extract the relevant task name from the ongoing scenario
        task_name = re.findall(r'Your task is to:\s*(.*?)\s*>', query_scenario)[2]
        # Find memories based on task name and temporal proximity
        similarity_results = self.scenario_memory.similarity_search_with_score(task_name, k=3)
        relevance_scores = []
        for result in similarity_results:
            trajectory = result[0].metadata['task_trajectory']
            prompt = f'''Evaluate how relevant this past task is to the current task based on the actions taken.Past Task:{trajectory}Current Task:{query_scenario}Your output format should be:Score: '''
            # Get importance score
            response = llm_response(prompt=prompt, model=self.llm_type, temperature=0.1, stop_strs=['\n'])
            score = int(re.search(r'\d+', response).group()) if re.search(r'\d+', response) else 0
            relevance_scores.append(score)
        # Select the most relevant memory based on the highest score
        if relevance_scores:
            max_score_idx = relevance_scores.index(max(relevance_scores))
            relevant_memory = similarity_results[max_score_idx][0].metadata['task_trajectory']
        return relevant_memory
    def addMemory(self, current_situation):
        # Extract task name
        task_name = re.search(r'Your task is to:\s*(.*?)\s*>', current_situation).group(1)
        # Summarize the critical actions taken in the task for better context
        summary_prompt = f'''Summarize the key actions taken to complete the following task:
{current_situation}
Your summary should be concise and focused on the actions taken. Summary: '''
        summary_response = llm_response(prompt=summary_prompt, model=self.llm_type, temperature=0.1)
        # Create document with metadata
        memory_doc = Document(
            page_content=summary_response,
            metadata={
                "task_name": task_name,
                "task_trajectory": current_situation})
        # Add to memory store
        self.scenario_memory.add_documents([memory_doc])

\end{lstlisting}
\end{tcolorbox}

\begin{tcolorbox}[colback=gray!10, colframe=black!50!black, title=The new agent found during the search on M3ToolEval]
\lstset{basicstyle=\scriptsize\ttfamily} 
\begin{lstlisting}
class WorkflowSelfRefine(workflowBase):
    def __call__(self, task_description: str, tool_instruction :str='', feedback :str= ''):
        prompt = '''Solve the task step by step
{task_description}'''
        prompt = prompt.format(task_description=task_description)
        workflow_result = llm_response(prompt=prompt, model=self.llm_type, temperature=0.1, stop_strs=['\n'])
        workflow_result = self.refine(task_description, worflow_result)
        return workflow_result
    def refine(self, task_description, workflow_result):
        ...
class MemoryEnhanced(MemoryBase):
    def retriveMemory(self, query_scenario): 
        # Return empty if no memories exist
        if self.scenario_memory._collection.count() == 0:
            return ''
        # Get top 3 similar memories
        similarity_results = self.scenario_memory.similarity_search_with_score(
            query_scenario, k=3) 
        fewshot_results = []
        importance_scores = []
        # Score each memory's relevance
        for result in similarity_results:
            trajectory = result[0].metadata['task_trajectory']
            fewshot_results.append(trajectory)
            # Generate prompt to evaluate importance
            prompt = f'''You will be given a successful case where you successfully complete the task. Then you will be given an ongoing task. Do not summarize these two cases, but rather evaluate how relevant and helpful the successful case is for the ongoing task, on a scale of 1-10.Success Case:{trajectory}Ongoing task:{query_scenario}Your output format should be:Score: '''
            # Get importance score
            response = llm_response(prompt=prompt, model=self.llm_type, temperature=0.1, stop_strs=['\n'])
            score = int(re.search(r'\d+', response).group()) if re.search(r'\d+', response) else 0
            importance_scores.append(score)
        # Return trajectory with highest importance score
        max_score_idx = importance_scores.index(max(importance_scores))
        return similarity_results[max_score_idx][0].metadata['task_trajectory']
    def addMemory(self, current_situation):
        # Extract task description
        task_name = current_situation.split('success.')[0]
        task_trajectory = current_situation.split('success.')[1]
        # Create document with metadata
        memory_doc = Document(
            page_content=task_name,
            metadata={
                "task_name": task_name, "task_trajectory": task_trajectory})
        # Add to memory store
        self.scenario_memory.add_documents([memory_doc])
class TooluseIO(TooluseBase):
    def __call__(self, task_description, tool_instruction, feedback_of_previous_tools):
        tool_pool = self.tooluse_pool.get(task_description)
        memory = self.get_memory(tool_instruction)
        prompt = f'''{memory}
You have access to the following tools:{tool_pool}
You need to select the appropriate tool from the list of available tools according to the task description to complete the task:{tool_instruction}
You must use the tools by outputing the tool name followed by its arguments, delimited by commas.
You can optionally express your thoughts using natural language before your action. For example, 'Thought: I want to use tool_name to do something. Action: <your action to call tool_name> End Action'.You can only invoke one tool at a time.
You must begin your tool invocation with 'Action:' and end it with 'End Action'.
Your tool invocation format must follow the invocation format in the tool description.
{feedback_of_previous_tools}'''        
        string = llm_response(prompt=prompt, model=self.llm_type, temperature=0.1) 
        return string

\end{lstlisting}
\end{tcolorbox}

\begin{tcolorbox}[colback=gray!10, colframe=black!50!black, title=The new agent found during the search on MATH]
\lstset{basicstyle=\scriptsize\ttfamily} 
\begin{lstlisting}
class WorflowCOTSC(WorflowBase):
    def __call__(self, task_description: str, feedback :str= ''):
        task_description = self.process_task_description(task_description)
        prompt = '''{memory}
Solve the problem step by step.
Here is the problem:
{task_description}'''
        prompt = prompt.format(task_description=task_description, memory=self.memory_cache)
        worflow_results = llm_response(prompt=prompt, model=self.llm_type, temperature=0.7, n=5)
        prompt = '''Below are several possible answers to a math problem. Please identify the answer that appears most frequently.
{worflow_results}

Please analyze each solution carefully and determine which answer is chosen by the majority. Return the most consistent answer.
'''
        prompt = prompt.format(worflow_results=worflow_results)
        worflow_result = llm_response(prompt=prompt, model=self.llm_type, temperature=0.0)
        return worflow_result
class AdaptiveHierarchicalPlanning(PlanningBase):
    def create_prompt(self, task_type, task_description, feedback):
        prompt = '''You are a planner who is an expert at creating step-by-step solution plans for {task_type} math problems.
Break down the problem into a minimal set of logical steps that lead to the solution.
Each step should be clear, focused on a specific mathematical operation, and described concisely.
Ensure the steps follow a coherent sequence where each builds upon previous results.
Task: {task_description}
'''
        prompt = prompt.format(task_description=task_description, task_type=task_type)
class AdaptiveMemory(MemoryBase):
    def retriveMemory(self, query_scenario):
        # Extract task name from query
        task_name = query_scenario
        # Return empty if no memories exist
        if self.scenario_memory._collection.count() == 0:
            return ''
        # Find most similar memories
        similarity_results = self.scenario_memory.similarity_search_with_score(task_name, k=2)
        # Extract trajectories from results
        memory_trajectories = [result[0].metadata['task_trajectory'] 
                             for result in similarity_results]               
        return '\n'.join(memory_trajectories)
    def addMemory(self, current_situation):
        # Prompt template for summarizing trajectory
        voyager_prompt = '''You are a helpful assistant that writes a description of the task resolution trajectory.
        1) Try to summarize the trajectory in no more than 6 sentences.
        2) Your response should be a single line of text.
        Trajectory:
        '''
        # Generate summarized trajectory
        prompt = voyager_prompt + current_situation
        trajectory_summary = llm_response(prompt=prompt, model=self.llm_type, temperature=0.1)
        # Create document with metadata
        doc = Document(
            page_content=trajectory_summary,
            metadata={
                "task_description": trajectory_summary,
                "task_trajectory": current_situation
            }
        )
        # Add to memory store
        self.scenario_memory.add_documents([doc])

\end{lstlisting}
\end{tcolorbox}
\end{document}